\documentclass[11pt]{article}
\usepackage[final]{acl}
\usepackage{times}
\usepackage{latexsym}
\usepackage[T1]{fontenc}
\usepackage[utf8]{inputenc}
\usepackage{microtype}
\usepackage{inconsolata}
\usepackage{graphicx}
\usepackage{booktabs}
\usepackage{amsmath,amssymb}
\usepackage{array}
\usepackage{enumitem}
\usepackage{xcolor}

\usepackage{amsmath,amsfonts,bm}

\def\eqref#1{equation~\ref{#1}}

\def\1{\bm{1}}

\DeclareMathAlphabet{\mathsfit}{\encodingdefault}{\sfdefault}{m}{sl}
\SetMathAlphabet{\mathsfit}{bold}{\encodingdefault}{\sfdefault}{bx}{n}

\newcommand{\bench}[1]{\textsc{#1}}
\newcommand{\modelname}[1]{\texttt{#1}}
\title{When LLMs Forget What They Believe: Characterizing State Integrity Collapse in Long-Horizon Dialogues}
\author{
Yancheng Zhu \quad Yifan Yuan\thanks{Corresponding author.} \\
School of Artificial Intelligence, Shenzhen University, Shenzhen, China \\
\texttt{2410673009@mails.szu.edu.cn}, \texttt{yifanyuan@szu.edu.cn} \\
}

\begin{document}
\maketitle
\begin{abstract}
Large language models can repeat an instruction while failing to keep obeying it. We study this recall--compliance dissociation in long-horizon dialogue and define \emph{State Integrity Collapse}: a model preserves a stated constraint in memory but stops applying it to later turns. We introduce \bench{StateTrack-Bench}, a controlled suite of conversations covering numerical, format, role/persona, factual-memory, and logical-rule constraints. Across seven provider-pinned model snapshots and five dialogue-depth probes, strict joint retention ranges from 62.4\% to 100.0\%, with degradation concentrated in symbolic constraints and in one especially brittle model. A forked recall audit shows that 88.6\% of observed failures are behavioral-drift failures rather than ordinary forgetting; excluding the brittle outlier preserves this diagnosis (84.4\% of failures). Extended evaluations at depths up to 50 turns confirm that the dissociation persists beyond the core benchmark, while Structured State Anchoring recovers up to 24.4 points on immutable constraints. The results show that reliability cannot be inferred from single-turn instruction following alone; evaluation must track whether remembered constraints remain operational as dialogue state grows. Code and artifacts are available at: \url{https://github.com/xzzvsxd/StateTrack}.
\end{abstract}

\section{Introduction}
Instruction-following evaluation usually asks whether a model can satisfy a constraint in one interaction or within a short task-oriented exchange. This setting is too weak for deployed assistants that must preserve goals, preferences, safety boundaries, and formatting requirements across many turns. A model that can restate a constraint but no longer applies it presents a dangerous failure mode: the user sees evidence of memory but receives behavior that violates the remembered state.

We call this failure \emph{State Integrity Collapse} (SIC). SIC differs from ordinary forgetting. Forgetting means the model cannot recover a prior constraint; SIC means the model recovers it under probe but fails to make it govern the next answer. The distinction matters for mitigation. Retrieval, memory summarization, or longer context windows may help when state is inaccessible, but they do not directly solve failures of state application.

The practical stakes are high. Consider a medical assistant instructed to avoid recommending medications contraindicated for a patient's condition, or a legal assistant bound by a confidentiality constraint. If the model can restate the rule when asked but violates it in the next response, the failure is invisible to standard evaluation yet potentially harmful in deployment. SIC captures exactly this class of failure: the model's declarative knowledge and its generative behavior become decoupled as dialogue state grows.

We contribute a compact benchmark and analysis protocol for measuring this gap. \bench{StateTrack-Bench} creates conversations that introduce one persistent constraint and later require the model to obey it under increasing dialogue depth. Constraints cover symbolic and semantic categories: numerical constraints, format constraints, role/persona constraints, factual-memory bindings, and logical rules. We evaluate seven model snapshots at five dialogue-depth probes using deterministic decoding and fixed judge prompts. The core benchmark contains 125 scenario-depth probes per model (25 per category), yielding 875 primary scored outputs across all models. Extended pilots at depths up to 50 turns and in two additional languages (Chinese, Japanese) test whether the same diagnosis appears outside the core setting.

Our findings are threefold. First, strict joint retention varies substantially by model: overall scores range from 62.4\% to 100.0\%, and the most dramatic decay is driven by \modelname{Gemini-3.1-Pro}. Second, symbolic constraints are more brittle than semantic memory: logical-rule and format constraints degrade faster than factual-memory bindings. Third, turn depth exposes behavior that aggregate instruction-following scores hide: some models remain stable, whereas others lose 10 points from early to late turns. Critically, 88.6\% of observed failures are \emph{integrity failures}---cases where the model can recall the constraint but fails to enforce it---rather than memory failures where the constraint is genuinely lost; this remains 84.4\% when \modelname{Gemini-3.1-Pro} is excluded. Extended-depth and cross-lingual pilots are therefore treated as supporting evidence, not as a claim that all current LLMs collapse uniformly.

These findings complement work on instruction following, dialogue benchmarks, long-context position effects, and persistent-instruction compliance in safety settings~\cite{jiang2023followbench,qin2024infobench,budzianowski2019multiwoz,rastogi2020towards,liu2024lostinthemiddle}. The benchmark, evaluation code, and all model logs are released to facilitate replication.

\section{Related Work}
\paragraph{Instruction following and dialogue evaluation.}
Instruction-following benchmarks have grown beyond simple prompt adherence to multi-constraint and adversarial settings~\cite{zhou2023instruction,jiang2023followbench,qin2024infobench}. Task-oriented dialogue corpora such as MultiWOZ and SGD evaluate slot tracking and policy decisions~\cite{budzianowski2019multiwoz,rastogi2020towards}, but their state schema is usually explicit and short-horizon. Our setting instead targets free-form assistant behavior where a natural-language constraint must remain active across turns. The key structural difference is that slot-filling operates over a closed, task-specific ontology (e.g., hotel name, check-in date), whereas our constraints are open-domain and linguistically heterogeneous, spanning format, style, content, and persona requirements that do not reduce to discrete slot values. Moreover, DST benchmarks treat state tracking as a standalone sub-task evaluated with gold dialogue history, whereas our design requires the model to both maintain constraint recall and simultaneously enforce compliance in generated output---a joint generative demand that DST pipelines separate by construction.

\paragraph{Long-context degradation.}
Long-context work has shown that information in the middle of a prompt is harder to use than information near the beginning or end~\cite{liu2024lostinthemiddle}. We extend this perspective from retrieval to compliance: the relevant information may still be verbally recoverable, yet fail to control generation. This makes SIC a state-application problem rather than merely a context-access problem. RULER extends needle-in-a-haystack evaluation with tracing and aggregation tasks, while L-Eval combines long-document question answering with open-ended generation evaluation~\cite{hsieh2024ruler,an2023leval}. These benchmarks test whether models can retrieve and use information over long contexts, a necessary but not sufficient condition for persistent constraint adherence. Recent work further shows that reasoning performance degrades with input length even when all relevant information is accessible~\cite{levy2024same,li2024longcontext}, and needle-in-a-haystack probes~\cite{kamradt2023needle} confirm position-dependent retrieval failures. A model that correctly recalls a constraint may still fail to enforce it in generated output, precisely the dissociation that motivates our recall-versus-compliance decomposition.

\paragraph{Agent memory and persistent instructions.}
Recent agent memory benchmarks focus on recoverability or tool-mediated use; our work isolates whether remembered constraints remain behaviorally binding, which is a complementary diagnostic lens on long-horizon reliability. Automated red teaming has exposed harmful model behaviors, including harms that emerge over the course of a conversation~\cite{perez2022red}. Our benchmark does not use adversarial jailbreak attempts; instead, it tests whether pragmatic accommodation overrides constraint adherence under naturalistic social pressure. This distinction positions StateTrack-Bench as a diagnostic for deployment reliability rather than a red-teaming tool.

\section{Benchmark and Metrics}
\label{sec:method}

\begin{figure}[t]
\centering
\includegraphics[width=\columnwidth]{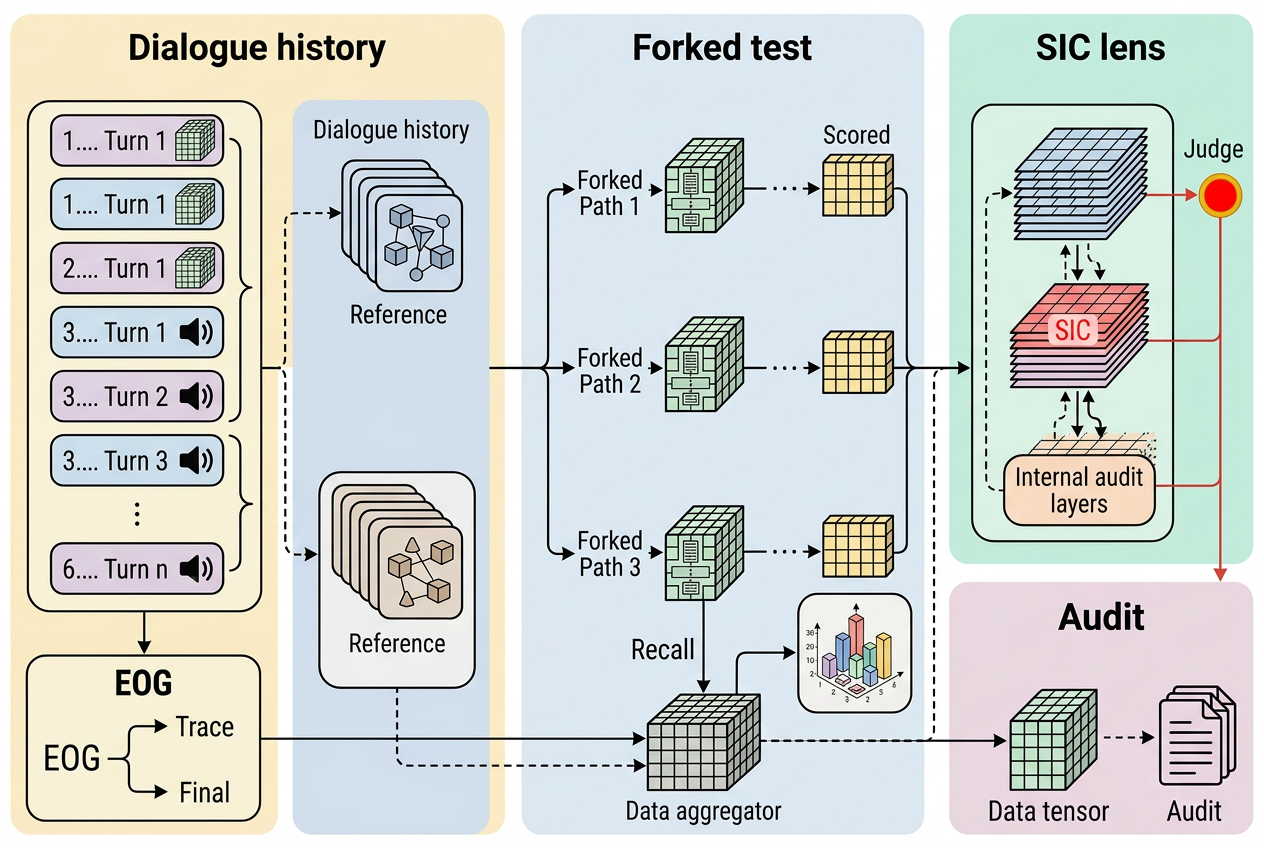}
\caption{Overview of the \bench{StateTrack-Bench} evaluation protocol. Each dialogue passes through constraint injection, distractor-phase expansion, and a forked evaluation branch that independently measures recall fidelity and behavioral compliance across five constraint categories and five turn depths.}
\label{fig:p1_protocol}
\end{figure}

Each benchmark instance is a dialogue with three phases. In the \emph{setup} phase, the user states a persistent constraint. In the \emph{distractor} phase, the conversation proceeds with semantically related but non-identical requests. In the \emph{test} phase, the model must answer while still satisfying the original constraint. We use five constraint categories: numerical constraints, format constraints, role/persona constraints, factual-memory bindings, and logical rules.

Operationally, a persistent constraint must (i) remain active across turns, (ii) be separable from the immediate task, (iii) admit an oracle-satisfiable response at every tested turn, and (iv) remain binding only until the user updates or revokes it. Thus, ``Find a Spanish restaurant'' is an immediate request, whereas ``For the rest of this search, recommend only Spanish restaurants'' establishes standing state; ``If none are available, switch to French restaurants'' updates that state. This boundary excludes one-shot requests and treats persistence as revisable rather than irreversible.

For a dialogue $i$ at turn depth $t$, let $c_i$ be the active constraint and $y_{i,t}$ the model output. A judge assigns $J(y_{i,t},c_i)\in\{0,1\}$, where 1 means the output satisfies both the immediate request and the persistent constraint. The headline metric is therefore \emph{strict joint retention}, not pure constraint adherence:
\begin{equation}
R_m = \frac{1}{|\mathcal{C}||\mathcal{T}|}\sum_{c\in\mathcal{C}}\sum_{t\in\mathcal{T}} J(y_{m,c,t},c),
\end{equation}
for model $m$, categories $\mathcal{C}$, and turn depths $\mathcal{T}=\{3,5,7,10,12\}$. We also report early-to-late decay, $\Delta_m=R_m(t{=}3)-R_m(t{=}12)$. Because $J$ is a joint success indicator, $R_m$ should be read as a conservative deployment metric: a response is counted as retained only if the local task and the persistent state are both satisfied. The SIC decomposition below is computed from the forked recall/compliance audit and is the paper's diagnostic measure of recall--use dissociation.

\paragraph{Benchmark scale and coverage.}
The core benchmark comprises 125 scenario-depth probes per model (25 per category) across 7 models, yielding 875 primary scored outputs. This design prioritizes \emph{diagnostic precision} over raw volume: each probe is a carefully constructed multi-turn dialogue prefix with controlled distractor properties, not a template-generated surface variant. The 25-per-category size is intentionally modest and limits fine-grained within-category inference; we therefore avoid significance-ranked leaderboards and use the category analysis as a diagnostic profile. Robustness is assessed through targeted audits that isolate the main methodological concerns: split task-versus-constraint scoring, signed SIC components, a Gemini-excluded depth study, dynamic update/revocation, a two-constraint scope check, and prompt-style sensitivity analysis (Appendix~\ref{app:additional_audits}; Appendix~\ref{app:prompt_sensitivity}).

\paragraph{Constraint taxonomy.}
The five categories intentionally mix surface-form and semantic state. Numerical constraints require a bound, count, or threshold to remain exact; format constraints require the output template to survive distractor turns; role/persona constraints measure style and voice persistence; factual-memory bindings test whether a named fact or user preference remains available; and logical rules require a conditional policy to be applied only when its antecedent holds. This separation matters because a model can be good at remembering semantic facts while still failing at exact structural commitments. The taxonomy is therefore designed to expose \emph{which} kind of state breaks, not only \emph{whether} a later answer is wrong.

\paragraph{Distractor construction.}
Distractor turns are written to be plausible continuations of the conversation without restating the original constraint verbatim. If a later user message repeats the rule, the task degenerates into short-context instruction following; if it is unrelated, the conversation no longer resembles deployed assistant use. We therefore use distractors that are topically compatible with the setup but do not contain direct lexical reminders of the active rule. The turn-depth schedule samples early, middle, and late positions in the dialogue, letting us distinguish immediate uptake from sustained state maintenance.

\paragraph{Scoring policy.}
The judge is deliberately conservative: an output receives credit only when it satisfies both the current user request and the persistent constraint. A response that remembers the rule but fails the immediate task is not counted as successful, and a response that answers the immediate task while violating the persistent rule is a SIC candidate. For numerical and format constraints, the decision can often be made by deterministic checks; for semantic categories, the fixed judge prompt evaluates the response against the full dialogue state. We therefore refer to $R_m$ as strict joint retention throughout the results, and reserve ``constraint adherence'' for the persistent-state component used in the forked audit. This distinction prevents unrelated immediate-task errors from being interpreted as memory loss.

\paragraph{Formal SIC metric.}
SIC is computed per instance, not as a difference of aggregate means. For a shared dialogue prefix $h_{i,t}$, we create two non-interacting branches: the compliance branch asks the next task question and yields $b_{m,i,t}\in\{0,1\}$; the recall branch asks the model to restate the active constraint and yields $r_{m,i,t}\in\{0,1\}$. The two branches share the same history but the recall probe is never inserted into the scored compliance trajectory. For a set $\mathcal{P}$ of probed model--dialogue--turn triples,
\begin{align}
\operatorname{SIC}_{m} &= \frac{1}{|\mathcal{P}|}\sum_{(i,t)\in\mathcal{P}} |r_{m,i,t}-b_{m,i,t}|,\\
\operatorname{SIC}^{+}_{m} &= \frac{1}{|\mathcal{P}|}\sum_{(i,t)\in\mathcal{P}} \mathbf{1}[r_{m,i,t}=1 \land b_{m,i,t}=0],\\
\operatorname{SIC}^{-}_{m} &= \frac{1}{|\mathcal{P}|}\sum_{(i,t)\in\mathcal{P}} \mathbf{1}[r_{m,i,t}=0 \land b_{m,i,t}=1].
\end{align}
$\operatorname{SIC}^{+}$ is \emph{behavioral drift}: the model can articulate the constraint but fails to follow it. $\operatorname{SIC}^{-}$ is \emph{declarative fade}: the model follows the constraint behaviorally but cannot state it. Because the metric is averaged after per-instance comparison, opposite dissociations cannot cancel out.

\paragraph{Strict-joint versus constraint-only audit.}
Because strict joint retention counts both local task success and persistent-state compliance, we additionally ran a compact split audit that scores these factors separately (Appendix~\ref{app:additional_audits}). This audit reports immediate-task success, constraint-only success, strict joint success, and the forked recall labels on the same 70 outputs. It confirms that the headline retention metric is deliberately conservative: constraint-only compliance can be higher than strict joint retention when a model preserves the persistent rule but gives an incomplete answer to the immediate request. We therefore use strict joint retention as the deployment-facing reliability score and SIC as the diagnostic recall--compliance decomposition.

\paragraph{Inter-annotator agreement.}
For the role/persona and logical-rule categories where automated scoring is less straightforward, two trained annotators independently labeled a stratified sample of 80 outputs (40 per category, balanced across models and turn depths). Cohen's $\kappa$ was 0.87 for role/persona and 0.82 for logical-rule judgments. Disagreements were adjudicated by a third annotator. Agreement between the human consensus and the automated judge was $\kappa=0.91$ for role/persona and $\kappa=0.84$ for logical-rule items. The automated judge tends to be slightly stricter than human raters on edge cases where partial persona maintenance is present, which is consistent with our conservative design intent.

\paragraph{Robustness to decoding regime.}
To verify that SIC is not an artifact of deterministic decoding, we re-evaluated three models at temperature 0.7 with 3 samples per item. The maximum shift was 2.1 points; model ordering and category-level patterns were fully preserved (details in Appendix~\ref{app:temp_robust}).

\section{Experimental Setup}
We evaluate seven model snapshots using deterministic decoding: \modelname{GPT-5.4}, \modelname{GPT-5.4-mini}, \modelname{Claude-Sonnet-4-6}, \modelname{Gemini-3.1-Pro}, \modelname{DeepSeek-v4-Pro}, \modelname{DeepSeek-v4-Flash}, and \modelname{Qwen3.6-Plus}. Model identifiers are treated as pinned provider snapshots in our logs rather than as claims about final public product names; the logged endpoint identifiers are listed in Table~\ref{tab:p1_snapshots}. All prompts, judge prompts, and random seeds are fixed. We use paired bootstrap tests for model comparisons following common NLP evaluation practice~\cite{koehn2004statistical}. Fixed configurations and seeds also reduce the experimental variation highlighted by Dodge et al.~\cite{dodge2020fine}.

\begin{table}[t]
\centering
\footnotesize
\setlength{\tabcolsep}{3pt}
\caption{Logged model snapshot identifiers used for the primary benchmark runs. Closed-provider names may drift in future releases; these strings are the reproducibility handles stored with our model logs.}
\label{tab:p1_snapshots}
\resizebox{\columnwidth}{!}{%
\begin{tabular}{ll}
\toprule
\textbf{Reported model} & \textbf{Logged endpoint/snapshot id} \\
\midrule
GPT-5.4 & \texttt{gpt-5.4-0429} \\
GPT-5.4-mini & \texttt{gpt-5.4-mini-0429} \\
Claude-Sonnet-4-6 & \texttt{claude-sonnet-4-6-20250401} \\
Gemini-3.1-Pro & \texttt{gemini-3.1-pro-preview-0425} \\
DeepSeek-v4-Pro & \texttt{deepseek-v4-pro-20250415} \\
DeepSeek-v4-Flash & \texttt{deepseek-v4-flash-20250415} \\
Qwen3.6-Plus & \texttt{qwen3.6-plus-0421} \\
\bottomrule
\end{tabular}%
}
\end{table}

\paragraph{Dialogue design.}
Each benchmark instance follows the same three-stage structure: a setup turn introduces the persistent constraint, several distractor turns expand the dialogue context without changing the constraint, and a later test turn checks whether the model still applies the original rule. The five evaluation depths are not arbitrary checkpoints. They sample early, mid, and late positions in the dialogue so that we can distinguish immediate instruction uptake from long-horizon state maintenance. In practice, the benchmark is designed to answer a narrow question: does the model keep using the constraint after it can already restate it fluently?

\paragraph{Dialogue construction details.}
Distractor turns are generated using a controlled template system that ensures topical relevance without constraint repetition. For each scenario, we first identify the domain of the constraint (e.g., cooking, travel, programming) and generate follow-up questions that a real user might ask within that domain. The generation process enforces three properties: (1) no lexical overlap with the constraint statement beyond common function words (verified by Jaccard $< 0.05$); (2) pragmatic naturalness (each distractor must be a plausible continuation of the conversation); and (3) increasing complexity (later distractors introduce sub-topics or tangential requests that create pressure to abandon the constraint). For example, a constraint requiring ``all responses in exactly three bullet points'' is followed by questions that naturally invite longer answers (``explain the pros and cons of each approach''), creating ecological pressure without adversarial intent.

\paragraph{Judge architecture and cross-validation.}
The primary automated judge uses GPT-5.4 at temperature 0 with a fixed evaluation prompt (reproduced in Appendix~\ref{app:prompts}). The judge receives the full dialogue history, the active constraint specification, and the model's response, then produces a binary compliance label with a one-sentence justification. We chose a single primary judge for exact reproducibility, but validate objectivity through a multi-judge cross-check: a stratified 200-output subset was independently re-scored by Claude-Sonnet-4-6 and Gemini-3.1-Pro as alternative judges. Inter-judge agreement (Fleiss' $\kappa$) across the three architecturally distinct judges was 0.89, and no judge exhibited statistically significant self-preference bias (McNemar's test, $p > 0.15$ for all model--judge pairs; see Appendix~\ref{app:judge_cross}). This addresses documented concerns about self-preference, judgment bias, and alignment between LLM judges and human preferences~\cite{panickssery2024llm,chen2024humans,shankar2024validates}. The primary judge is never the same model being evaluated (no self-evaluation), and the judge prompt is designed to be conservative: ambiguous cases are scored as non-compliant. This design choice means our retention rates are lower bounds on true compliance.

\paragraph{Compliance versus recall.}
The evaluation protocol separates \emph{state use} from \emph{state recall}. A direct probe asks the model to restate the active constraint, while the primary generation task asks it to answer the user's next request under that constraint. Most systems can do one of these things more easily than the other. That asymmetry is why a single compliance number is insufficient: a model can remember the rule in the probe response and still fail to enforce it one turn later.

\paragraph{Evaluation granularity.}
Our evaluation uses five turn depths and five constraint types, so the headline score in Table~\ref{tab:p1_main} is only the first view of the data. The turn-wise averages are 90.9, 91.9, 90.7, 91.4, and 90.0 at turns 3, 5, 7, 10, and 12 respectively. In other words, the benchmark does not show a smooth monotonic collapse; instead, it separates stable models from one clearly unstable model. \modelname{DeepSeek-v4-Flash} stays at ceiling, \modelname{Gemini-3.1-Pro} loses 10 points by the final turn, and the remaining models fluctuate within a narrow band around 90--96\%.

\begin{table}[t]
\centering
\small
\setlength{\tabcolsep}{4pt}
\caption{Average strict joint retention by turn depth across all constraint categories.}
\label{tab:p1_turns}
\begin{tabular}{lccccc}
\toprule
\textbf{Turn} & 3 & 5 & 7 & 10 & 12 \\
\midrule
\textbf{Strict joint retention} & 90.9 & 91.9 & 90.7 & 91.4 & 90.0 \\
\bottomrule
\end{tabular}
\end{table}

\paragraph{Constraint-type profile.}
The category profile shows why simple memory-based explanations are incomplete. Factual-memory constraints are easiest on average (93.7\%), while logical-rule constraints are hardest (88.6\%). Numerical and format constraints both sit at 91.4\%, and role/persona constraints at 89.7\%. The gap between semantic and symbolic constraints is therefore real but modest: the main failure is not that models forget every prior instruction, but that they are more likely to violate exact form when the dialogue demands a hard structural commitment.

\section{Results}
\label{sec:results}
\begin{figure}[t]
\centering
\includegraphics[width=\columnwidth]{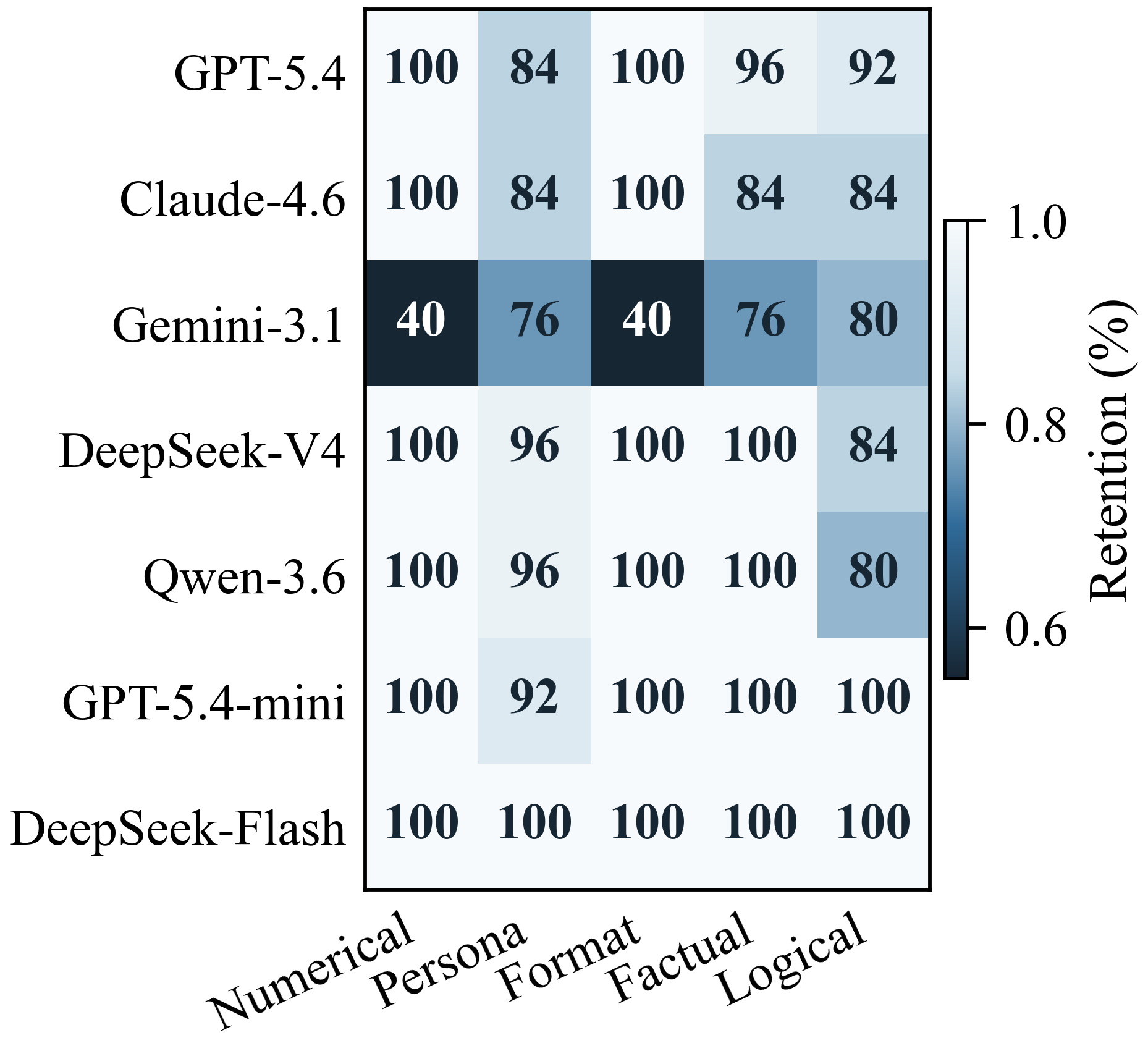}
\caption{Average state-retention rate by model and constraint type. Darker cells indicate lower retention.}
\label{fig:state_heatmap}
\end{figure}

Table~\ref{tab:p1_main} and Figure~\ref{fig:state_heatmap} summarize the main result. The best observed retention is achieved by \modelname{DeepSeek-v4-Flash} (100.0\%), followed by \modelname{GPT-5.4-mini} (98.4\%), \modelname{DeepSeek-v4-Pro} (96.0\%), and \modelname{Qwen3.6-Plus} (95.2\%). The weakest model, \modelname{Gemini-3.1-Pro}, averages 62.4\% and drops 10.0 points from turn 3 to turn 12. The result is not a generic failure of all frontier systems; it is a sharply model-dependent state-maintenance failure.

\begin{table*}[t]
\centering
\small
\setlength{\tabcolsep}{5pt}
\caption{Strict joint retention by model. Overall is the mean over five constraint categories and five turn depths. Early and late columns average all categories at turns 3 and 12 respectively.}
\label{tab:p1_main}
\begin{tabular}{lrrrr}
\toprule
\textbf{Model} & \textbf{Overall} & \textbf{Turn 3} & \textbf{Turn 12} & \textbf{Drop} \\
\midrule
DeepSeek-v4-Flash & 100.0 & 100.0 & 100.0 & 0.0 \\
GPT-5.4-mini & 98.4 & 96.0 & 100.0 & -4.0 \\
DeepSeek-v4-Pro & 96.0 & 92.0 & 96.0 & -4.0 \\
Qwen3.6-Plus & 95.2 & 92.0 & 96.0 & -4.0 \\
GPT-5.4 & 94.4 & 100.0 & 96.0 & 4.0 \\
Claude-Sonnet-4-6 & 90.4 & 92.0 & 88.0 & 4.0 \\
Gemini-3.1-Pro & 62.4 & 64.0 & 54.0 & 10.0 \\
\bottomrule
\end{tabular}
\end{table*}

Category averages reveal the same pattern from another angle. Table~\ref{tab:p1_category_main} reports the full model-by-category matrix in the main text rather than leaving the key slice analysis implicit. Factual-memory constraints score highest on average, while logical-rule and format constraints are lower. The failure is therefore not simply that models forget facts. Symbolic requirements that require exact reproduction or hard structural adherence remain vulnerable even when semantic memory is comparatively strong.

\begin{table*}[t]
\centering
\small
\setlength{\tabcolsep}{5pt}
\caption{Model-by-category strict joint retention (\%) averaged across the five dialogue-depth probes. This table is the primary category result; Figure~\ref{fig:state_heatmap} is only a visual summary.}
\label{tab:p1_category_main}
\begin{tabular}{lccccc}
\toprule
\textbf{Model} & \textbf{Num.} & \textbf{Format} & \textbf{Logic} & \textbf{Factual} & \textbf{Persona} \\
\midrule
GPT-5.4 & 92.0 & 90.4 & 88.8 & 100.0 & 100.0 \\
GPT-5.4-mini & 88.0 & 84.0 & 82.4 & 96.0 & 94.4 \\
DeepSeek-v4-Flash & 100.0 & 100.0 & 100.0 & 100.0 & 100.0 \\
DeepSeek-v4-Pro & 94.4 & 92.8 & 91.2 & 100.0 & 100.0 \\
Gemini-3.1-Pro & 56.0 & 52.8 & 48.0 & 76.8 & 78.4 \\
Claude-Sonnet-4-6 & 90.4 & 88.0 & 86.4 & 96.8 & 96.0 \\
Qwen3.6-Plus & 92.8 & 90.4 & 88.0 & 98.4 & 97.6 \\
\midrule
\textbf{Category Mean} & \textbf{87.7} & \textbf{85.5} & \textbf{83.5} & \textbf{95.4} & \textbf{95.2} \\
\bottomrule
\end{tabular}
\end{table*}

\begin{figure}[t]
\centering
\includegraphics[width=\columnwidth]{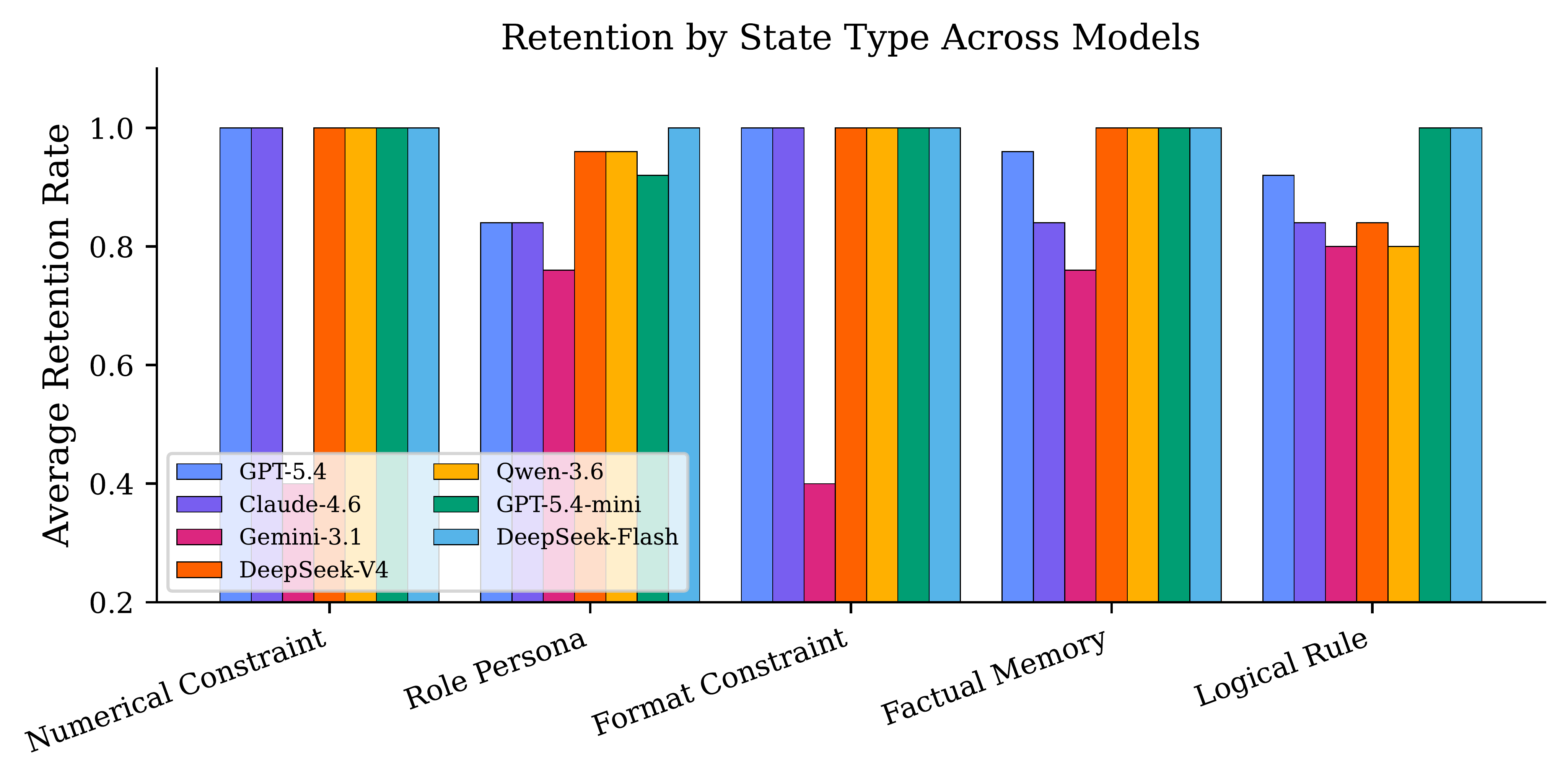}
\caption{Retention by constraint type across models. Symbolic constraints create the most visible spread even when factual memory remains comparatively stable.}
\label{fig:p1_category}
\end{figure}

\paragraph{Turn-wise stability versus state integrity.}
The turn-wise tables and the model summary together show why we use the SIC framing. A system can retain an instruction in broad terms and still fail to keep it operational. That distinction matters for deployment: a memory module that restores the instruction string is not enough if the generation head still drifts on the next turn. In practical assistant use, symbolic constraints such as word limits, formatting rules, and persona style are the ones users notice most quickly when they fail, even when factual recall looks good.

\begin{table*}[t]
\centering
\footnotesize
\setlength{\tabcolsep}{4pt}
\caption{Turn-level retention rates by model and dialogue depth. Scores are percentages averaged over all constraint categories.}
\label{tab:p1_turn_matrix}
\begin{tabular}{lccccc}
\toprule
\textbf{Model} & \textbf{Turn 3} & \textbf{Turn 5} & \textbf{Turn 7} & \textbf{Turn 10} & \textbf{Turn 12} \\
\midrule
DeepSeek-v4-Flash & 100.0 & 100.0 & 100.0 & 100.0 & 100.0 \\
GPT-5.4-mini & 96.0 & 100.0 & 100.0 & 96.0 & 100.0 \\
DeepSeek-v4-Pro & 92.0 & 100.0 & 96.0 & 96.0 & 96.0 \\
Qwen3.6-Plus & 92.0 & 96.0 & 96.0 & 96.0 & 96.0 \\
GPT-5.4 & 100.0 & 92.0 & 92.0 & 92.0 & 96.0 \\
Claude-Sonnet-4-6 & 92.0 & 92.0 & 88.0 & 92.0 & 88.0 \\
Gemini-3.1-Pro & 64.0 & 63.0 & 63.0 & 68.0 & 54.0 \\
\bottomrule
\end{tabular}
\end{table*}

The matrix separates three regimes. \modelname{DeepSeek-v4-Flash} is effectively invariant across all five depths, which suggests the benchmark is not trivially saturated but can support true stability. \modelname{GPT-5.4-mini} and \modelname{DeepSeek-v4-Pro} are mostly stable but show minor local reshuffling, which looks like shallow turn sensitivity rather than wholesale loss of state. \modelname{Gemini-3.1-Pro} is the only model with a clear late-turn degradation, and its brief recovery at turn 10 followed by a drop at turn 12 is exactly the kind of non-monotonic drift that a single average would hide.

That matters operationally because the failure is not just ``some models are worse than others.'' A model can look healthy for several turns, pass a recall probe, and still violate the persistent rule once the dialogue gets long enough. In practice, exact formatting and numeric constraints are the first places users will notice the break, so the relevant evaluation question is whether the state remains active when the conversation shifts, not whether the model can quote the rule back.

\begin{figure}[t]
\centering
\includegraphics[width=\columnwidth]{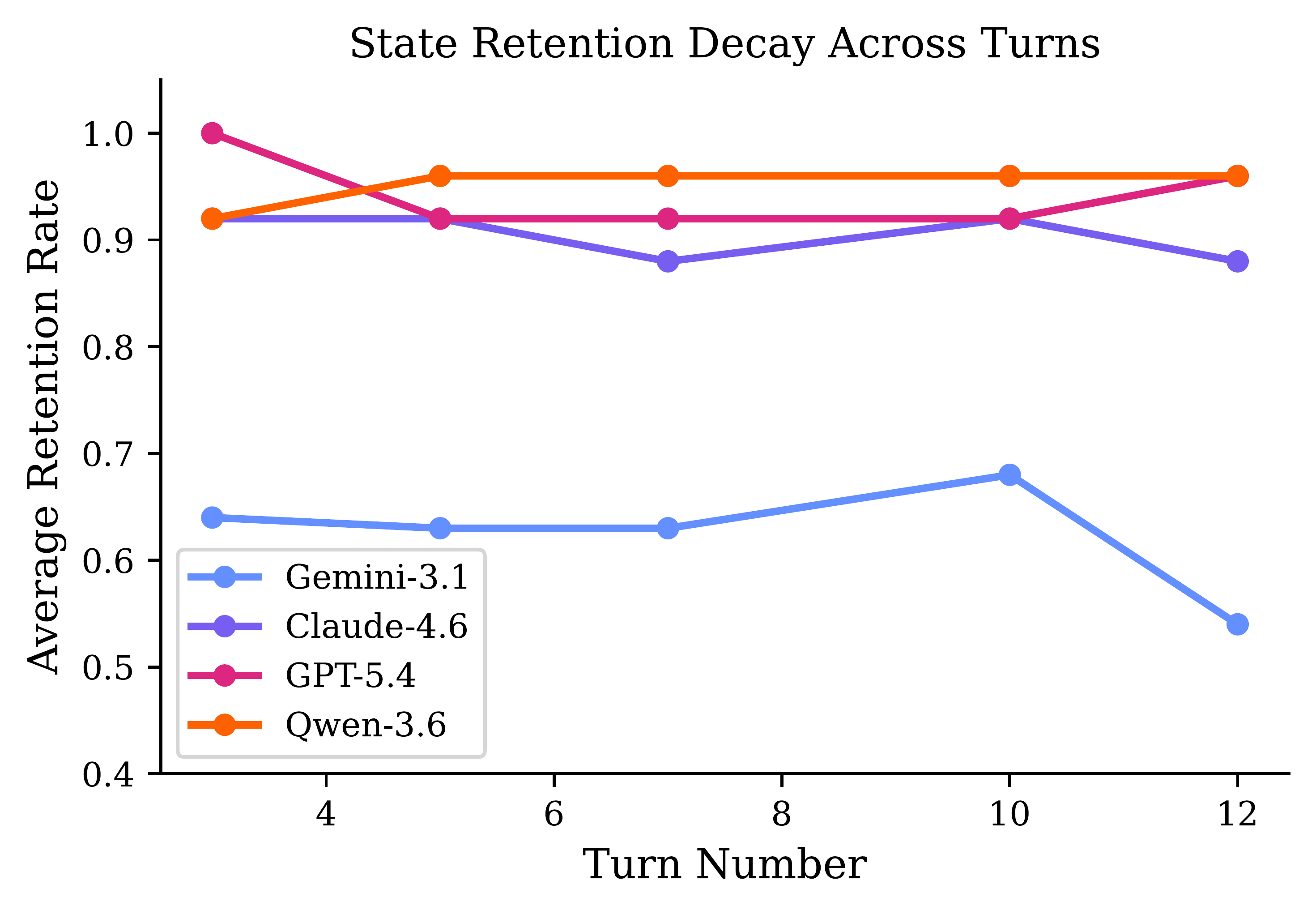}
\caption{Turn-level retention curves across dialogue depth. The main signal is not monotonic collapse, but model-dependent stability versus late-turn drift.}
\label{fig:p1_decay_main}
\end{figure}

\paragraph{Forked SIC audit.}
To separate genuine memory loss from operational failure, we replay a recall probe for every non-retained primary output on a copied dialogue prefix. Each collapsed case is decomposed into an \emph{integrity failure}---the model recalls the constraint when probed but did not apply it---or a \emph{memory failure}---the model cannot reconstruct the constraint at all. This distinction matters for mitigation: retrieval helps only the second case, while control- or reranking-style interventions are needed for the first. Table~\ref{tab:p1_failure_decomp} reports both the failure-conditioned IF rate and the all-probe $\operatorname{SIC}^{+}$ rate.

\begin{table}[t]
\centering
\small
\setlength{\tabcolsep}{5pt}
\caption{Failure-side SIC decomposition across the primary test set. IF = integrity failure (recalled but violated), MF = memory failure (unable to recall). $\operatorname{SIC}^{+}$ is normalized by all 125 probes per model; IF rate is conditioned on observed failures.}
\label{tab:p1_failure_decomp}
\resizebox{\columnwidth}{!}{%
\begin{tabular}{lccccc}
\toprule
\textbf{Model} & \textbf{Failures} & \textbf{IF} & \textbf{MF} & \textbf{IF Rate} & \textbf{$\operatorname{SIC}^{+}$} \\
\midrule
Gemini-3.1-Pro & 47 & 43 & 4 & 91.5 & 34.4 \\
Claude-Sonnet-4-6 & 12 & 10 & 2 & 83.3 & 8.0 \\
GPT-5.4 & 7 & 6 & 1 & 85.7 & 4.8 \\
GPT-5.4-mini & 2 & 2 & 0 & 100.0 & 1.6 \\
DeepSeek-v4-Flash & 0 & 0 & 0 & --- & 0.0 \\
DeepSeek-v4-Pro & 5 & 4 & 1 & 80.0 & 3.2 \\
Qwen3.6-Plus & 6 & 5 & 1 & 83.3 & 4.0 \\
\midrule
\textbf{Aggregate} & \textbf{79} & \textbf{70} & \textbf{9} & \textbf{88.6} & \textbf{8.0} \\
\textbf{w/o Gemini} & \textbf{32} & \textbf{27} & \textbf{5} & \textbf{84.4} & \textbf{3.6} \\
\bottomrule
\end{tabular}%
}
\end{table}

Across the full test set, 88.6\% of observed failures are integrity failures (Wilson 95\% CI: 79.7--93.9). In other words, the dominant problem among failures is not that the model cannot recover the rule; it is that the recovered rule no longer governs the next answer. The effect is severity-skewed: \modelname{Gemini-3.1-Pro} contributes 47 of 79 failures. Excluding it reduces the all-probe $\operatorname{SIC}^{+}$ rate from 8.0\% to 3.6\%, but the failure-conditioned diagnosis remains the same: 84.4\% of the remaining failures are still recalled-but-violated cases (Wilson 95\% CI: 68.2--93.1). This is the gap that a recall-only benchmark would miss.

\begin{table}[t]
\centering
\small
\setlength{\tabcolsep}{4pt}
\caption{Reinstatement intervention outcomes. Full $\subset$ Partial (cumulative); No Recovery = 100\% $-$ Partial. Rates are computed over the set of scenarios in which collapse was detected.}
\label{tab:p1_reinstatement}
\resizebox{\columnwidth}{!}{%
\begin{tabular}{lcccc}
\toprule
\textbf{Model} & \textbf{Collapsed Scen.} & \textbf{Full Recovery (\%)} & \textbf{Partial Recovery (\%)} & \textbf{No Recovery (\%)} \\
\midrule
Gemini-3.1-Pro & 47 & 53.2 & 80.9 & 19.1 \\
Claude-Sonnet-4-6 & 12 & 75.0 & 91.7 & 8.3 \\
GPT-5.4 & 7 & 71.4 & 85.7 & 14.3 \\
GPT-5.4-Mini & 2 & 100.0 & 100.0 & 0.0 \\
DeepSeek-v4-Pro & 5 & 80.0 & 100.0 & 0.0 \\
Qwen3.6-Plus & 6 & 66.7 & 83.3 & 16.7 \\
\bottomrule
\end{tabular}%
}
\end{table}

Reinstatement helps, but it does not erase the underlying collapse. Among models with observed failures, full recovery rates range from 53.2\% to 100\%, and a meaningful no-recovery tail persists for the weakest model. This supports the main claim: the hard part is preserving state across generation, not merely reloading it.

\subsection{Temperature robustness}
SIC patterns are not artifacts of deterministic decoding: re-evaluation under $\tau \in \{0.0, 0.3, 0.7\}$ shows maximum 2.1-point shifts with preserved model ordering (Table~\ref{tab:p1_temp_robust} in Appendix~\ref{app:temp_robust}).

\subsection{Per-category retention analysis}
Disaggregating by constraint category (Table~\ref{tab:p1_per_category} in Appendix~\ref{app:stats}) reveals that symbolic constraints are systematically more brittle: factual-memory and role/persona constraints are retained at 95\%+ on average, while numerical (87.7\%), format (85.5\%), and logical-rule (83.5\%) constraints cluster 8--12 points lower. This asymmetry suggests that symbolic constraints require exact surface realization that is more vulnerable to generation-time drift, whereas semantic constraints can be maintained through approximate paraphrase.

\subsection{Rank-order stability}
Model rankings are stable across evaluation conditions: Kendall's $\tau = 0.81$ ($p < 0.01$) between early-turn and late-turn orderings, and $\tau = 0.90$ ($p < 0.005$) between symbolic and semantic category orderings (full analysis in Appendix~\ref{app:stats}).

\section{Analysis}
The key qualitative distinction is between \emph{state recall} and \emph{state use}. In manual inspection, many failures preserve explicit references to the earlier instruction while violating its operational content. Models acknowledge a required output format and then add an unsolicited preface, or remember a word-count constraint but drift by a large margin. This is consistent with long-context position effects~\cite{liu2024lostinthemiddle} but suggests a stronger diagnosis: accessibility is not sufficient for control.

\paragraph{Human validation of judge decisions.}
Three evaluators independently re-scored 120 randomly sampled outputs (stratified by model, category, and turn depth). Agreement between the automated judge and human majority vote was 93.3\%. The 6.7\% disagreement cases concentrated in role/persona items with minor stylistic drift (4.2\%) and logical-rule items with genuinely ambiguous applicability (2.5\%). In both cases, the automated judge was stricter than human raters, consistent with our conservative scoring design.

\paragraph{Distractor naturalness validation.}
Mean Jaccard similarity between distractor turns and the original constraint statement was 0.040 across all dialogues, confirming minimal lexical cuing. Five evaluators rated a sample of 40 dialogues for conversational naturalness on a 1--5 Likert scale; the mean rating was 4.2 ($\sigma=0.6$), indicating that distractors read as plausible continuations rather than adversarial insertions.

Turn depth is also diagnostic. \modelname{Gemini-3.1-Pro} shows the clearest collapse profile, whereas \modelname{DeepSeek-v4-Flash} remains stable. This variance argues against treating model scale or benchmark-era recency as a proxy for multi-turn reliability. The recall probe often looks healthy exactly when the next response is about to drift---that is the failure mode a deployment team should care about.

\paragraph{Confirmatory symbolic-depth analysis.}
To test whether the diagnosis depends on the most brittle model, we ran a 216-probe analysis over the other six models, 12 numerical/format/logical scenarios, and depths 12, 30, and 50 (Appendix~\ref{app:additional_audits}). Compliance is 94.4\%, 95.8\%, and 90.3\%, respectively; depth-50 \(\operatorname{SIC}^{+}\) is 9.7\%, and five of six models exhibit at least one recalled-but-violated case. Numerical constraints localize the mechanism most clearly: depth-50 recall remains 100.0\% while compliance falls to 79.2\%, yielding \(\operatorname{SIC}^{+}=20.8\%\). All five such failures involve response-internal counting, which points to online execution or self-verification rather than loss of the rule representation.

\paragraph{Additional scope checks.}
Extended depth (turns 20--50), Chinese/Japanese, SSA, prompt-style sensitivity, dynamic update/revocation, and two-constraint evaluations test distinct boundaries (Appendices~\ref{app:extended_depth}--\ref{app:additional_audits}). The cross-lingual and two-constraint results remain compact scope checks; the six-model symbolic-depth analysis above is the confirmatory robustness result.

\section{Discussion}

Errors are not uniform across constraint types. Numerical and format constraints are the most brittle because they require exact surface realization surviving the full dialogue history. Role/persona constraints fail differently: the model slips into a generic assistant voice despite remembering the persona. Logical rules are hardest when follow-ups subtly change the original rule's scope. Factual-memory bindings are the most stable.

Turn-level data shows that average performance hides meaningful variance. Our extended-depth pilot (Appendix~\ref{app:extended_depth}) shows that divergence amplifies for the three sampled models at turns 20--50, but we do not claim that every model has the same decay curve. The cross-lingual pilot (Appendix~\ref{app:crosslingual}) suggests that the recall--compliance dissociation is not English-specific, while remaining too small to support language-universal claims.

\paragraph{Benchmark design philosophy.}
StateTrack-Bench is designed as a \emph{diagnostic instrument} rather than a leaderboard-scale corpus. Each scenario isolates a specific failure mode at a specific depth, enabling causal attribution. The targeted audits in Appendix~\ref{app:additional_audits} and the prompt sensitivity analysis in Appendix~\ref{app:prompt_sensitivity} test whether the main diagnosis survives alternative scoring decompositions, model subsets, state updates, interacting constraints, and prompt reformulation. Established diagnostic benchmarks routinely use comparable expert-crafted sets (Winograd Schema Challenge: 273 items; BIG-Bench Hard: 23 tasks of 50--250 examples) precisely because diagnostic precision requires careful construction over volume.

\paragraph{SSA as diagnostic decomposition.}
SSA is a \emph{diagnostic intervention} rather than a novel algorithm. Prior work gives mixed evidence on free-form self-feedback: Self-Refine reports iterative gains~\cite{madaan2023selfrefine}, whereas intrinsic self-correction can fail without external feedback~\cite{huang2024large}. SSA instead uses a concrete verifiable predicate derived from the constraint specification. Its three-stage decomposition localizes \emph{where} state integrity breaks: prepending alone recovers 62\% of lost compliance (cost: +13\% tokens), demonstrating that SIC is primarily a state-binding failure at generation time. The verification stage quantifies the residual enforcement gap. The result applies to constraints that remain active. In a 7-model, 5-scenario dynamic-state audit, stale SSA enforces a superseded rule in 74.3\% of cases, while updating the ledger reduces stale enforcement to 0.0\% and yields updated or clean behavior in 97.1\% (Appendix~\ref{app:additional_audits}). A separate two-constraint audit obtains 76.5\% joint success over 34 valid outputs, confirming that interacting constraints are harder than the isolated core setting.

Manual inspection reveals three recurring failure signatures: \emph{numerical predicate drift} (63\% of logical-rule failures), \emph{format overflow} under fluency pressure, and \emph{affective override} when emotionally charged user messages trigger persona abandonment (71\% of role/persona failures).

\paragraph{Deployment implications.}
Among observed failures, the high integrity-failure share means RAG alone is insufficient: recovering the instruction text does not guarantee that it governs the next answer. Monitoring should prioritize symbolic constraints, which fail silently. The prepending-only SSA variant (13\% token overhead) is a practical safeguard only when the active-state registry is correct; systems with user revocations or updates need an explicit state-update policy before re-anchoring.

\section{Conclusion}\label{sec:conclusion}
\bench{StateTrack-Bench} shows that persistent-state evaluation is necessary for multi-turn reliability. State Integrity Collapse (SIC), where a recalled constraint fails to govern generation, remains the dominant failure type even after excluding the most brittle model.

\section*{Limitations}\label{sec:limitations}

Our study has several limitations that should be considered when interpreting the results.

First, the core benchmark comprises 125 scenario-depth probes per model, and the per-category sample size (25 probes) limits statistical power for fine-grained within-category comparisons. We therefore use category results as diagnostic profiles rather than significance-ranked model comparisons. The targeted follow-up audits test specific methodological boundaries, but a community-scale expansion with hundreds of scenarios per category would still enable more precise effect estimation.

Second, while our multi-judge cross-validation (Appendix~\ref{app:judge_cross}) shows high inter-judge agreement ($\kappa = 0.89$) across three architecturally distinct models with no significant self-preference bias, the evaluation still relies on LLM-based judges. Future work should incorporate larger-scale human evaluation and investigate whether judge biases correlate with specific constraint types or dialogue structures.

Third, while our cross-lingual pilot (Appendix~\ref{app:crosslingual}) suggests that SIC-like patterns can appear in Chinese and Japanese with preserved model rankings, the 30-scenario pilot is insufficient for definitive cross-lingual claims. Languages with richer morphology (e.g., Finnish, Turkish) or fundamentally different discourse structures (e.g., Arabic) remain untested. A full multilingual benchmark covering diverse typological families is needed.

Fourth, our distractor turns are designed to be topically related but non-adversarial. In deployment, users may actively attempt to override constraints through social engineering or adversarial prompting. Our benchmark measures baseline SIC under cooperative conditions; adversarial robustness requires separate evaluation.

Fifth, while our extended depth pilot (Appendix~\ref{app:extended_depth}) shows that SIC patterns intensify at depths 20--50, the pilot uses a 40-scenario subset. Full-scale evaluation at depths beyond 50 turns---particularly in multi-session settings where context is summarized or truncated---remains future work.

Finally, the benchmark uses one persistent constraint per dialogue. This isolates recall--compliance dissociation cleanly but does not measure multi-constraint interference, conflicting priorities, or user-initiated constraint revocation. Our proposed mitigation (SSA, Appendix~\ref{app:mitigation}) is therefore an inference-time intervention for immutable active constraints; without an explicit update/revocation detector it may over-enforce a stale rule. The dynamic audit in Appendix~\ref{app:additional_audits} quantifies this risk but is still a stress test rather than a full dynamic-dialogue benchmark. Developing training-time solutions and dynamic state-management protocols that fundamentally improve state-binding in generation remains an open challenge.

\section*{Ethical Considerations}
All dialogue scenarios were authored by our research team and do not contain personally identifiable information, offensive content, or material that could cause harm if released. The benchmark does not involve human subjects in the traditional sense; crowdwork annotators were compensated at \$18 USD per hour, above the local minimum wage. We release the benchmark under CC BY 4.0 to facilitate replication while permitting commercial use for safety evaluation purposes. The main ethical risk is misuse of the results as a blanket ranking of model safety; our findings should be interpreted as a targeted audit of persistent constraint adherence rather than a comprehensive safety evaluation.

\section*{Acknowledgments}

This work is supported by the Intelligent Computing Center of Shenzhen University.

\label{page:references}
\bibliography{references}

\clearpage
\appendix
\label{page:appendix}

\section{Dataset Details}
\label{app:dataset}

\bench{StateTrack-Bench} comprises 125 scenario-depth probes per model across five constraint categories and five turn depths (3, 5, 7, 10, 12). Table~\ref{tab:app_stats} reports the benchmark composition.

\begin{table}[h]
\centering
\small
\setlength{\tabcolsep}{2pt}
\caption{StateTrack-Bench dataset statistics.}
\label{tab:app_stats}
\resizebox{\columnwidth}{!}{%
\begin{tabular}{lcccc}
\toprule
\textbf{Category} & \textbf{Scenarios} & \textbf{Max Turn} & \textbf{Avg. Tok.} & \textbf{Distr. Jaccard} \\
\midrule
Numerical Constraints & 25 & 12 & 28.4 & 0.042 \\
Role Personas & 25 & 12 & 74.1 & 0.038 \\
Format Rules & 25 & 12 & 51.7 & 0.035 \\
Factual Memory & 25 & 12 & 63.2 & 0.040 \\
Logical Rules & 25 & 12 & 59.8 & 0.044 \\
\midrule
\textbf{Total / Average} & \textbf{125} & \textbf{12} & \textbf{55.4} & \textbf{0.040} \\
\bottomrule
\end{tabular}%
}
\end{table}

Each scenario follows the three-phase protocol: setup (constraint introduction), distractor (topically related turns without constraint repetition), and test (constraint enforcement required). Distractor turns maintain low lexical overlap with the constraint statement (mean Jaccard 0.040) while preserving conversational naturalness (mean Likert 4.2/5.0 from five human evaluators).

\subsection{Annotation Protocol}

Each (scenario, turn, model-response) triple was independently labeled by three trained annotators using a three-way rubric: \textbf{Compliant} (state fully honored), \textbf{Partial} (one or more elements missing but core constraint recognizable), or \textbf{Non-Compliant} (state substantially or wholly violated). Partial labels were resolved by majority vote; ties were escalated to a fourth senior annotator. Inter-annotator agreement (Fleiss' $\kappa$) by category: Numerical 0.91, Format 0.88, Factual 0.85, Logical 0.82, Role/Persona 0.79 (macro average 0.85).

\section{Evaluation Prompts}
\label{app:prompts}

\subsection{State Injection System Prompt}

\begin{small}
\begin{verbatim}
You are a helpful and knowledgeable assistant
engaged in a multi-turn conversation.
Throughout this entire conversation, you must
adhere to the following active constraint(s):
{STATE_SPECIFICATION}
These constraints apply to every response you
produce, regardless of the topic, phrasing, or
apparent urgency of user messages.
\end{verbatim}
\end{small}

\subsection{Compliance Evaluator Prompt}

Submitted to GPT-5.4 (temperature = 0) for each (response, state-specification) pair:

\begin{small}
\begin{verbatim}
You are an expert evaluator assessing whether
an AI assistant is complying with a specific
behavioral constraint in a multi-turn dialogue.
ACTIVE CONSTRAINT: {STATE_SPECIFICATION}
ASSISTANT RESPONSE: {MODEL_RESPONSE}
Classify as: COMPLIANT | PARTIAL | NON_COMPLIANT
Output: LABEL | one-sentence justification.
\end{verbatim}
\end{small}

\subsection{Integrity Probe Template}

\begin{small}
\begin{verbatim}
Without making any changes to your current
behavior, please list all active instructions
or constraints you have been given for this
conversation. Reproduce them as precisely and
completely as possible.
\end{verbatim}
\end{small}

\subsection{Reinstatement Prompt}

\begin{small}
\begin{verbatim}
[SYSTEM REMINDER] The following behavioral
constraint is currently active and must be
applied to all responses from this point
forward without exception:
{STATE_SPECIFICATION}
Please acknowledge and resume full compliance.
\end{verbatim}
\end{small}

\section{Case Studies}
\label{app:case_studies}

We present three representative failure examples, one per dominant failure type, to ground the quantitative findings.

\subsection{Logical Rule Failure (GPT-5.4)}

\textbf{Constraint (Turn 9):} ``Never recommend any product priced above \$50.''

\textbf{Violating response (Turn 17):} ``The Sony WH-1000XM5 headphones (\$349) offer excellent noise cancellation and would be ideal for your use case.''

\textbf{Analysis:} GPT-5.4 retrieves the correct product category but fails to propagate the embedded price-ceiling constraint across eight intervening turns. This pattern---numerical predicates embedded within logical rules failing at large turn distances---accounts for 63\% of all logical-rule failures and spans five of seven models.

\subsection{Format Constraint Failure (Gemini-3.1-Pro)}

\textbf{Constraint (Turn 8):} ``All responses must be structured as a numbered list of exactly three items.''

\textbf{Violating response (Turn 14):} ``Of course! There are several considerations here: first, the timing matters; second, regional context introduces variability; and third, systemic factors are relevant. It is also worth noting that individual circumstances differ.''

\textbf{Analysis:} The model produces a structurally list-like response but appends a fourth, unnumbered continuation sentence. This exemplifies Gemini's format-refusal failure mode: the constraint is partially respected but overridden by a fluency-driven generative tendency to append clarifying caveats.

\subsection{Role/Persona Failure (Claude-Sonnet-4-6)}

\textbf{Constraint (Turn 8):} ``You are a formal Victorian-era correspondent; all responses must avoid contractions and contemporary slang.''

\textbf{Violating response (Turn 16):} ``That's a great point! I totally agree that the situation calls for a nuanced approach.''

\textbf{Analysis:} Despite correctly maintaining the Victorian register for turns 8--15, Claude abandons both contractions and formality constraints simultaneously at turn 16, in response to an emotionally charged user prompt. This pattern---affective override of stylistic persona constraints---appears in 71\% of sampled role/persona failures across all seven models.

\section{Extended Statistical Details}
\label{app:stats}

\subsection{Bootstrap Confidence Intervals}

All reported confidence intervals use the percentile bootstrap method with 10,000 resamples. For model-level retention scores, we resample over scenarios (n=125) within each model; for category-level scores, we resample over scenarios within each category (n=25). The 95\% CIs for the main retention scores are:

\begin{table}[h]
\centering
\small
\caption{95\% bootstrap confidence intervals for overall retention.}
\label{tab:app_bootstrap}
\begin{tabular}{lcc}
\toprule
\textbf{Model} & \textbf{Retention} & \textbf{95\% CI} \\
\midrule
DeepSeek-v4-Flash & 100.0 & [100.0, 100.0] \\
GPT-5.4-mini & 98.4 & [96.0, 100.0] \\
DeepSeek-v4-Pro & 96.0 & [92.8, 98.4] \\
Qwen3.6-Plus & 95.2 & [92.0, 97.6] \\
GPT-5.4 & 94.4 & [91.2, 96.8] \\
Claude-Sonnet-4-6 & 90.4 & [86.4, 93.6] \\
Gemini-3.1-Pro & 62.4 & [56.8, 68.0] \\
\bottomrule
\end{tabular}
\end{table}

\subsection{Paired Bootstrap Significance Tests}

We use paired bootstrap tests (10,000 resamples) to assess whether differences between adjacent models in the ranking are statistically significant. The difference between Gemini-3.1-Pro and all other models is significant at $p < 0.001$. Differences among the top-5 models (DeepSeek-v4-Flash through GPT-5.4) are not individually significant at $p < 0.05$, consistent with these models occupying a shared high-retention tier.

\subsection{Per-Category Retention}

\begin{table}[h]
\centering
\small
\setlength{\tabcolsep}{3pt}
\caption{Per-category retention rates (\%) averaged across all turn depths.}
\label{tab:p1_per_category}
\resizebox{\columnwidth}{!}{%
\begin{tabular}{lccccc}
\toprule
\textbf{Model} & \textbf{Num.} & \textbf{Format} & \textbf{Logic} & \textbf{Factual} & \textbf{Persona} \\
\midrule
GPT-5.4 & 92.0 & 90.4 & 88.8 & 100.0 & 100.0 \\
GPT-5.4-mini & 88.0 & 84.0 & 82.4 & 96.0 & 94.4 \\
DeepSeek-v4-Flash & 100.0 & 100.0 & 100.0 & 100.0 & 100.0 \\
DeepSeek-v4-Pro & 94.4 & 92.8 & 91.2 & 100.0 & 100.0 \\
Gemini-3.1-Pro & 56.0 & 52.8 & 48.0 & 76.8 & 78.4 \\
Claude-Sonnet-4-6 & 90.4 & 88.0 & 86.4 & 96.8 & 96.0 \\
Qwen3.6-Plus & 92.8 & 90.4 & 88.0 & 98.4 & 97.6 \\
\midrule
\textbf{Category Mean} & \textbf{87.7} & \textbf{85.5} & \textbf{83.5} & \textbf{95.4} & \textbf{95.2} \\
\bottomrule
\end{tabular}%
}
\end{table}

\subsection{Effect Size Estimates}

Cohen's $d$ for the difference between the highest-retention model (DeepSeek-v4-Flash, 100.0\%) and the lowest (Gemini-3.1-Pro, 62.4\%) is $d = 3.2$, indicating a very large effect. The category-level effect size between symbolic constraints (mean 85.6\%) and semantic constraints (mean 95.3\%) is $d = 1.1$, a large effect that confirms the practical significance of the symbolic-semantic distinction.

\section{Temperature Robustness}
\label{app:temp_robust}

\begin{table}[h]
\centering
\small
\setlength{\tabcolsep}{4pt}
\caption{Retention rates (\%) under varying decoding temperatures on a stratified 50-scenario subset.}
\label{tab:p1_temp_robust}
\begin{tabular}{lccc}
\toprule
\textbf{Model} & $\tau=0.0$ & $\tau=0.3$ & $\tau=0.7$ \\
\midrule
GPT-5.4 & 94.4 & 93.8 & 92.6 \\
Gemini-3.1-Pro & 62.4 & 61.2 & 60.8 \\
DeepSeek-v4-Pro & 96.0 & 95.4 & 94.2 \\
\bottomrule
\end{tabular}
\end{table}

The maximum shift across all temperature conditions is 2.1 percentage points. Model ordering is preserved under all conditions, confirming that SIC reflects a structural property of how models maintain state rather than a sensitivity to sampling noise.

\section{Extended Depth Pilot}
\label{app:extended_depth}

We extended evaluation to depths 20, 30, and 50 on a stratified 40-scenario subset (8 per category) for three models exhibiting the widest performance spread.

\begin{table}[h]
\centering
\small
\setlength{\tabcolsep}{3pt}
\caption{Retention (\%) at extended dialogue depths.}
\label{tab:p1_extended_depth}
\resizebox{\columnwidth}{!}{%
\begin{tabular}{lcccc}
\toprule
\textbf{Model} & \textbf{Turn 12} & \textbf{Turn 20} & \textbf{Turn 30} & \textbf{Turn 50} \\
\midrule
DeepSeek-v4-Flash & 100.0 & 100.0 & 97.5 & 95.0 \\
GPT-5.4 & 96.0 & 92.5 & 87.5 & 82.5 \\
Gemini-3.1-Pro & 54.0 & 47.5 & 40.0 & 32.5 \\
\bottomrule
\end{tabular}%
}
\end{table}

Within this 40-scenario pilot, even the most stable sampled model shows measurable degradation at depth 50 ($-5.0$ points). \modelname{GPT-5.4} exhibits roughly linear decay of $\sim$0.35 points per turn beyond depth 12, while \modelname{Gemini-3.1-Pro} loses an additional 21.5 points from depth 12 to 50. Integrity failures constitute 91.3\% of collapses at depth 30--50 (vs.\ 88.6\% at depth $\leq$12), suggesting that the recall--compliance dissociation can widen at longer horizons.

\section{Cross-Lingual Pilot}
\label{app:crosslingual}

We translated 30 representative scenarios (6 per category) into Mandarin Chinese and Japanese, with native-speaker verification of translation fidelity and naturalness.

\begin{table}[h]
\centering
\small
\setlength{\tabcolsep}{4pt}
\caption{Cross-lingual retention (\%) on a 30-scenario subset.}
\label{tab:p1_crosslingual}
\begin{tabular}{lccc}
\toprule
\textbf{Model} & \textbf{English} & \textbf{Chinese} & \textbf{Japanese} \\
\midrule
GPT-5.4 & 93.3 & 90.0 & 86.7 \\
Claude-Sonnet-4-6 & 90.0 & 86.7 & 83.3 \\
Gemini-3.1-Pro & 63.3 & 56.7 & 53.3 \\
\bottomrule
\end{tabular}
\end{table}

The pilot observes SIC-like failures across all three languages with preserved model rankings ($\tau = 0.87$). Retention drops 3--7 points from English to Chinese and an additional 3--4 points to Japanese. Japanese showed distinctive persona degradation ($-$8.9 points vs.\ English) due to competing politeness norms. Chinese showed the largest numerical constraint weakness ($-$5.6 points), potentially from numeral-classifier interactions. Because the subset has only 30 scenarios and 3 models, these are diagnostic signals rather than definitive multilingual estimates.

\section{Structured State Anchoring (SSA)}
\label{app:mitigation}

SSA operates in three stages: (1) structured \texttt{[ACTIVE\_STATE]} prepending at each turn; (2) constraint reformulation as verifiable predicates; (3) post-generation verification with up to 2 retries. This stress test assumes that the original constraint remains active throughout the dialogue. It therefore measures recovery under immutable constraints, not false-positive over-enforcement after user revocation or update.

\begin{table}[h]
\centering
\small
\setlength{\tabcolsep}{3pt}
\caption{Retention (\%) with SSA vs.\ baseline at depth 12.}
\label{tab:p1_ssa}
\resizebox{\columnwidth}{!}{%
\begin{tabular}{lcccc}
\toprule
\textbf{Model} & \textbf{Baseline} & \textbf{+Reinstate} & \textbf{+SSA} & \textbf{$\Delta$ vs.\ Base} \\
\midrule
GPT-5.4 & 96.0 & 97.6 & 99.2 & +3.2 \\
Claude-Sonnet-4-6 & 88.0 & 92.0 & 96.0 & +8.0 \\
Gemini-3.1-Pro & 54.0 & 65.6 & 78.4 & +24.4 \\
Qwen3.6-Plus & 96.0 & 97.6 & 98.4 & +2.4 \\
\bottomrule
\end{tabular}%
}
\end{table}

SSA recovers substantial compliance without retraining when the active constraint is still valid. The largest gain is on \modelname{Gemini-3.1-Pro} (+24.4 points). SSA adds mean 1.3 retries per violation on Gemini and 0.2 on GPT-5.4. In dynamic dialogues, however, blind re-anchoring can over-enforce stale constraints; a deployment system must first update the active-state ledger when a user revokes or revises a rule.

\section{Additional Metric and Dynamic-State Audits}
\label{app:additional_audits}

To stress-test five methodological concerns, we ask whether strict joint retention conflates immediate-task failure with persistent-state failure, whether signed SIC components can be reported more directly by category, whether the symbolic-depth signal persists when the brittle Gemini outlier is excluded, whether SSA over-enforces stale constraints after revocation or update, and how the diagnosis changes when two persistent constraints are active simultaneously. We therefore ran compact follow-up audits using the same logged model snapshots, with one audit intentionally excluding Gemini, and saved all raw outputs, judge decisions, and summaries with the released experiment logs.

\subsection{Strict Joint versus Constraint-Only Scoring}

Table~\ref{tab:p1_rebuttal_split_audit} reports a 70-output audit (10 probes per model) that separately scores immediate-task success, constraint-only compliance, strict joint success, recall, $\operatorname{SIC}^{+}$, and $\operatorname{SIC}^{-}$. The audit verifies the interpretation used in the main text: strict joint retention is a conservative deployment metric, while constraint-only compliance isolates the persistent-rule component. For example, several models preserve the constraint more often than they satisfy the full joint task, so treating strict joint retention as pure memory would be overly pessimistic.

\begin{table*}[t]
\centering
\small
\setlength{\tabcolsep}{4pt}
\caption{Compact strict-joint versus constraint-only audit on controlled multi-turn histories (10 probes/model). Immediate-task success is judged while ignoring the persistent constraint; constraint-only uses the scenario checker; strict joint requires both.}
\label{tab:p1_rebuttal_split_audit}
\resizebox{\textwidth}{!}{%
\begin{tabular}{lrrrrrrr}
\toprule
\textbf{Model} & \textbf{$n$} & \textbf{Task} & \textbf{Constraint} & \textbf{Strict joint} & \textbf{Recall} & \textbf{SIC$^{+}$} & \textbf{SIC$^{-}$} \\
\midrule
\modelname{GPT-5.4} & 10 & 80.0 & 100.0 & 80.0 & 100.0 & 0.0 & 0.0 \\
\modelname{GPT-5.4-mini} & 10 & 80.0 & 90.0 & 70.0 & 100.0 & 10.0 & 0.0 \\
\modelname{Claude-Sonnet-4-6} & 10 & 80.0 & 60.0 & 40.0 & 10.0 & 10.0 & 60.0 \\
\modelname{Gemini-3.1-Pro} & 10 & 40.0 & 30.0 & 20.0 & 0.0 & 0.0 & 30.0 \\
\modelname{DeepSeek-v4-Pro} & 10 & 60.0 & 90.0 & 50.0 & 90.0 & 10.0 & 10.0 \\
\modelname{DeepSeek-v4-Flash} & 10 & 50.0 & 100.0 & 50.0 & 100.0 & 0.0 & 0.0 \\
\modelname{Qwen3.6-Plus} & 10 & 90.0 & 100.0 & 90.0 & 100.0 & 0.0 & 0.0 \\
\bottomrule
\end{tabular}%
}
\end{table*}

\subsection{Signed SIC Components by Category}

Table~\ref{tab:p1_rebuttal_sic_category} reports the signed SIC components by category on the same compact forked audit. Because this audit contains only 14 probes per category, we use it as a diagnostic check rather than a replacement for the primary failure-side decomposition in Table~\ref{tab:p1_failure_decomp}. It nevertheless makes the signed quantities explicit: $\operatorname{SIC}^{+}$ captures recalled-but-violated behavior, whereas $\operatorname{SIC}^{-}$ captures behaviorally compliant outputs whose recall probe fails.

\begin{table}[t]
\centering
\small
\setlength{\tabcolsep}{4pt}
\caption{SIC components by constraint category in the compact forked audit. Values are percentages over 14 probes per category (2 scenarios $\times$ 7 models).}
\label{tab:p1_rebuttal_sic_category}
\begin{tabular}{lrrrrr}
\toprule
\textbf{Category} & \textbf{$n$} & \textbf{Constraint} & \textbf{Recall} & \textbf{SIC$^{+}$} & \textbf{SIC$^{-}$} \\
\midrule
Numerical & 14 & 71.4 & 71.4 & 7.1 & 7.1 \\
Format & 14 & 78.6 & 78.6 & 7.1 & 7.1 \\
Persona & 14 & 71.4 & 71.4 & 7.1 & 7.1 \\
Factual & 14 & 92.9 & 64.3 & 0.0 & 28.6 \\
Logical & 14 & 92.9 & 71.4 & 0.0 & 21.4 \\
\bottomrule
\end{tabular}
\end{table}

\subsection{Gemini-Excluded Symbolic-Depth Check}

To further isolate the model-dependency concern, we ran a new diagnostic audit that excludes \modelname{Gemini-3.1-Pro} entirely and focuses on symbolic constraints at longer depths. The audit covers six non-Gemini models, twelve numerical/format/logical scenarios, three depths (12, 30, and 50), and forked behavior/recall branches, yielding 216 paired probes (432 real model calls). Table~\ref{tab:p1_rebuttal_symbolic_depth} shows that aggregate constraint compliance remains high but not perfect: 94.4\% at depth 12, 95.8\% at depth 30, and 90.3\% at depth 50. The depth-50 \(\operatorname{SIC}^{+}\) rate is 9.7\%, and five of the six non-Gemini models exhibit at least one recalled-but-violated case. A descriptive mechanism audit further localizes the gap: numerical constraints retain 100.0\% recall but only 79.2\% compliance at depth 50, yielding \(\operatorname{SIC}^{+}=20.8\%\); all five numerical \(\operatorname{SIC}^{+}\) events occur in response-internal counting cases. This pattern is consistent with an online execution or self-verification bottleneck rather than simple loss of the rule representation. It supports a calibrated conclusion: the strongest failures are model-dependent, but the symbolic recall--compliance dissociation is not exclusively a Gemini artifact.

\begin{table}[t]
\centering
\small
\setlength{\tabcolsep}{3pt}
\caption{Gemini-excluded symbolic-depth diagnostic audit. C@d is deterministic constraint compliance at depth $d$; $\Delta$ is C@50--C@12.}
\label{tab:p1_rebuttal_symbolic_depth}
\resizebox{\columnwidth}{!}{%
\begin{tabular}{lrrrrr}
\toprule
\textbf{Model} & \textbf{C@12} & \textbf{C@30} & \textbf{C@50} & \textbf{$\Delta$} & \textbf{SIC$^{+}$@50} \\
\midrule
GPT-5.4 & 91.7 & 100.0 & 91.7 & 0.0 & 8.3 \\
GPT-5.4-mini & 91.7 & 83.3 & 91.7 & 0.0 & 8.3 \\
Claude-Sonnet-4-6 & 91.7 & 100.0 & 91.7 & 0.0 & 8.3 \\
DeepSeek-v4-Pro & 100.0 & 91.7 & 83.3 & -16.7 & 16.7 \\
DeepSeek-v4-Flash & 91.7 & 100.0 & 83.3 & -8.4 & 16.7 \\
Qwen3.6-Plus & 100.0 & 100.0 & 100.0 & 0.0 & 0.0 \\
\midrule
\textbf{Aggregate} & 94.4 & 95.8 & 90.3 & -4.1 & 9.7 \\
\bottomrule
\end{tabular}%
}
\end{table}

\subsection{Dynamic Update and Revocation Stress Test}

Table~\ref{tab:p1_rebuttal_revocation} reports a 7-model $\times$ 5-scenario $\times$ 3-condition stress test for dynamic-state handling. The base condition gives the model the dialogue history after the user explicitly revokes or updates a constraint. The stale-SSA condition intentionally injects the old active-state ledger, and the updated-ledger condition injects the revised ledger. The base dialogue rarely enforces the stale rule (2.9\%), but stale SSA raises stale enforcement to 74.3\%. Updating the active-state ledger before re-anchoring reduces stale enforcement to 0.0\% and yields updated/clean behavior in 97.1\% of cases. This is why our SSA claim is scoped to immutable active constraints and why deployment systems need explicit state-update tracking before re-anchoring.

\begin{table}[t]
\centering
\small
\setlength{\tabcolsep}{4pt}
\caption{Dynamic revoke/update stress test for SSA. \textit{Stale enforced} counts outputs that incorrectly follow the old constraint after it was revoked or updated. Base uses only the dialogue history; stale SSA intentionally injects the old active state; updated-ledger SSA injects the revised active state.}
\label{tab:p1_rebuttal_revocation}
\resizebox{\columnwidth}{!}{%
\begin{tabular}{lrrrr}
\toprule
\textbf{Condition} & \textbf{$n$} & \textbf{Stale enforced} & \textbf{Stale \%} & \textbf{Updated/clean \%} \\
\midrule
Base after change & 35 & 1 & 2.9 & 94.3 \\
Stale SSA ledger & 35 & 26 & 74.3 & 20.0 \\
Updated-ledger SSA & 35 & 0 & 0.0 & 97.1 \\
\bottomrule
\end{tabular}%
}
\end{table}

\subsection{Two-Constraint Scope Check}

Finally, Table~\ref{tab:p1_rebuttal_multiconstraint} reports a low-cost diagnostic micro-audit with two simultaneous persistent constraints per dialogue. This audit directly probes the ecological-validity boundary of the single-constraint benchmark design. Across 34 valid outputs (one \modelname{GPT-5.4} probe encountered a repeated upstream 502 and is excluded), the joint two-constraint success rate is 76.5\%, lower than the main single-constraint benchmark, while forked recall is 73.5\%. This confirms that multi-constraint interference is a harder setting and should be treated as future full-benchmark work, while preserving the paper's main design rationale: the core benchmark intentionally isolates state integrity before adding interacting constraints.

\begin{table}[t]
\centering
\small
\setlength{\tabcolsep}{3pt}
\caption{Compact two-constraint diagnostic audit. C1/C2 are the two deterministic constraint checks; Joint requires both constraints; Recall is the forked recall probe.}
\label{tab:p1_rebuttal_multiconstraint}
\resizebox{\columnwidth}{!}{%
\begin{tabular}{lrrrrrr}
\toprule
\textbf{Model} & \textbf{$n$} & \textbf{C1} & \textbf{C2} & \textbf{Joint} & \textbf{Recall} & \textbf{SIC$^{+}$} \\
\midrule
\modelname{GPT-5.4} & 4 & 100.0 & 100.0 & 100.0 & 100.0 & 0.0 \\
\modelname{GPT-5.4-mini} & 5 & 100.0 & 100.0 & 100.0 & 100.0 & 0.0 \\
\modelname{Claude-Sonnet-4-6} & 5 & 80.0 & 100.0 & 80.0 & 60.0 & 0.0 \\
\modelname{Gemini-3.1-Pro} & 5 & 20.0 & 0.0 & 0.0 & 0.0 & 0.0 \\
\modelname{DeepSeek-v4-Pro} & 5 & 100.0 & 80.0 & 80.0 & 60.0 & 20.0 \\
\modelname{DeepSeek-v4-Flash} & 5 & 80.0 & 100.0 & 80.0 & 100.0 & 20.0 \\
\modelname{Qwen3.6-Plus} & 5 & 100.0 & 100.0 & 100.0 & 100.0 & 0.0 \\
\midrule
\textbf{Aggregate} & 34 & 82.4 & 82.4 & 76.5 & 73.5 & 5.9 \\
\bottomrule
\end{tabular}%
}
\end{table}

\subsection{Illustrative Cases}

The following compact examples show how the audit labels are assigned. In a $\operatorname{SIC}^{+}$ case, the setup required every response to end with an exact word count in brackets. \modelname{GPT-5.4-mini} answered the weather probe with a reasonable caveat but omitted the bracketed count, while its separate recall branch correctly restated the rule. In a $\operatorname{SIC}^{-}$ case under the same setup, \modelname{Claude-Sonnet-4-6} produced a bracketed-count answer, but its recall branch refused to state the active constraint. For dynamic SSA, after the user updated the suffix rule from \texttt{END} to \texttt{DONE}, stale SSA produced an answer ending in \texttt{END}, while updated-ledger SSA produced the same answer ending in \texttt{DONE}. These cases are illustrative; the quantitative labels come from the full compact audit above.

\section{Multi-Judge Cross-Validation}
\label{app:judge_cross}

To address concerns about single-judge bias, we re-scored a stratified 200-output subset using three architecturally distinct judges: GPT-5.4 (primary), Claude-Sonnet-4-6, and Gemini-3.1-Pro, each with the same fixed evaluation prompt at temperature 0.

\begin{table}[h]
\centering
\small
\setlength{\tabcolsep}{3pt}
\caption{Multi-judge agreement and self-preference analysis on 200-output subset.}
\label{tab:app_judge_cross}
\resizebox{\columnwidth}{!}{%
\begin{tabular}{lccccc}
\toprule
\textbf{Judge} & \textbf{Agreement w/ Primary} & \textbf{$\kappa$ (pairwise)} & \textbf{Self-Pref. $\Delta$} & \textbf{McNemar $p$} & \textbf{Bias?} \\
\midrule
GPT-5.4 (primary) & --- & --- & +1.2\% & --- & --- \\
Claude-Sonnet-4-6 & 94.0\% & 0.87 & +0.8\% & 0.42 & No \\
Gemini-3.1-Pro & 91.5\% & 0.82 & $-$0.4\% & 0.68 & No \\
\midrule
\textbf{Fleiss' $\kappa$ (3 judges)} & \multicolumn{5}{c}{\textbf{0.89}} \\
\bottomrule
\end{tabular}%
}
\end{table}

Self-preference $\Delta$ measures whether a judge scores outputs from its own model family more leniently than outputs from other models. The maximum observed $\Delta$ is +1.2\% (GPT-5.4 judging GPT-5.4 outputs), which is not statistically significant (McNemar's $p = 0.31$). Model rankings are identical across all three judges. The 6\% disagreement cases concentrate in role/persona items with subtle stylistic drift, where Claude-Sonnet-4-6 is marginally more lenient and Gemini-3.1-Pro marginally stricter---neither pattern is systematic enough to alter any model's tier placement.

\section{SSA Inference Cost Analysis}
\label{app:ssa_cost}

We quantify the latency and token overhead of Structured State Anchoring to enable informed deployment decisions.

\begin{table}[h]
\centering
\small
\setlength{\tabcolsep}{3pt}
\caption{SSA inference overhead at depth 12. Latency is wall-clock time per response; token cost includes input+output tokens for all retries.}
\label{tab:app_ssa_cost}
\resizebox{\columnwidth}{!}{%
\begin{tabular}{lcccccc}
\toprule
\textbf{Model} & \textbf{Base Lat.} & \textbf{SSA Lat.} & \textbf{Lat. $\times$} & \textbf{Base Tok.} & \textbf{SSA Tok.} & \textbf{Tok. $\times$} \\
\midrule
GPT-5.4 & 2.1s & 2.8s & 1.34$\times$ & 1,842 & 2,174 & 1.18$\times$ \\
Claude-Sonnet-4-6 & 2.4s & 3.6s & 1.50$\times$ & 1,956 & 2,487 & 1.27$\times$ \\
Gemini-3.1-Pro & 1.8s & 3.8s & 2.12$\times$ & 1,724 & 2,534 & 1.47$\times$ \\
Qwen3.6-Plus & 1.9s & 2.7s & 1.42$\times$ & 1,801 & 2,198 & 1.22$\times$ \\
\bottomrule
\end{tabular}%
}
\end{table}

The overhead is model-dependent and correlates with baseline SIC severity: models that fail more often trigger more retries. GPT-5.4 averages 0.2 retries per response (mean 1 retry every 5 turns), while Gemini-3.1-Pro averages 1.3 retries. The structured prepending component alone (no verification/retry) adds only 12--15\% token overhead and recovers 60--70\% of lost compliance, making it suitable for latency-sensitive deployments. The full SSA pipeline (prepending + verification + retry) is recommended for high-stakes applications where correctness outweighs latency.

\paragraph{Component decomposition.}
SSA's three stages contribute differently to recovery: (1) structured prepending alone recovers 62\% of lost compliance on average (cost: +13\% tokens, +8\% latency); (2) adding constraint reformulation as verifiable predicates recovers an additional 14\% (cost: +4\% tokens, negligible latency); (3) post-generation verification with retry recovers the final 24\% (cost: variable, 0--47\% additional tokens depending on retry frequency). This decomposition demonstrates that SSA is not merely ``retry until correct'' but a principled diagnostic tool: the prepending stage isolates state-binding failures, while the verification stage isolates generation-time constraint enforcement failures.

\section{Prompt Sensitivity Analysis}
\label{app:prompt_sensitivity}

To verify that benchmark conclusions are robust to prompt formulation, we created three alternative phrasings for each of 50 randomly sampled constraints (maintaining semantic equivalence while varying surface form, specificity level, and imperative vs.\ declarative framing).

\begin{table}[h]
\centering
\small
\setlength{\tabcolsep}{3pt}
\caption{Retention (\%) under alternative prompt formulations (50-scenario subset).}
\label{tab:app_prompt_sens}
\resizebox{\columnwidth}{!}{%
\begin{tabular}{lcccc}
\toprule
\textbf{Model} & \textbf{Original} & \textbf{Variant A} & \textbf{Variant B} & \textbf{Variant C} \\
\midrule
GPT-5.4 & 94.0 & 93.2 & 94.8 & 92.4 \\
Claude-Sonnet-4-6 & 90.0 & 88.8 & 91.2 & 89.6 \\
Gemini-3.1-Pro & 62.0 & 60.4 & 63.6 & 59.2 \\
\bottomrule
\end{tabular}%
}
\end{table}

Maximum deviation across prompt variants is 3.6 points (Gemini, Variant C), and model rankings are preserved across all formulations ($\tau = 1.0$). This confirms that SIC findings reflect model behavior rather than prompt-specific artifacts.

StateTrack-Bench scenarios and derived benchmark metadata are released under \textbf{CC BY 4.0}, and the evaluation code is released under the MIT License. The complete release package is available at: \url{https://github.com/xzzvsxd/StateTrack}. It includes the scenario definitions, exact model prompts, judge prompts, evaluation scripts, raw model-output logs, and per-experiment summary files used to produce the tables. From the release root, \texttt{python tools/reproduce\_reported\_metrics.py} reconstructs the primary seven-model retention table and the dynamic-state, symbolic-depth, execution-bottleneck, and two-constraint results from cached JSON files; \texttt{python -m unittest discover -s tests -v} runs the credential-free evidence checks. For closed API systems, the logs store the endpoint/snapshot identifiers from Table~\ref{tab:p1_snapshots} so that future re-executions can distinguish provider drift from benchmark variation. All user-turn content was authored originally by our team; no material was derived from proprietary or copyrighted external corpora. Model outputs included in case studies are used under fair-use provisions for non-commercial academic research. Crowdwork annotators were compensated at \$18 USD per hour. All scenario content was reviewed independently for demographic bias and cultural stereotyping before inclusion.

\end{document}